\documentclass[letterpaper, 10 pt, conference]{ieeeconf}

\IEEEoverridecommandlockouts
\usepackage{graphicx}
\usepackage{amsmath,amsfonts,amssymb}
\usepackage{xcolor}
\usepackage{siunitx}
\usepackage{booktabs}
\usepackage{multirow}
\let\labelindent\relax
\usepackage{enumitem}

\usepackage{tikz}
\usetikzlibrary{positioning, fit, backgrounds, calc, arrows.meta, decorations.pathreplacing, shapes.geometric}

\usepackage{hyperref}
\usepackage{cleveref}
\crefname{figure}{Fig.}{Figs.}
\Crefname{figure}{Fig.}{Figs.}
\crefname{table}{Table}{Tables}
\Crefname{table}{Table}{Tables}

\usepackage[style=ieee, citestyle=numeric-comp, maxcitenames=2, mincitenames=1]{biblatex}
\AtBeginBibliography{\footnotesize}

\newif\ifcenterfigcaptions
\makeatletter
\long\def\@makecaption#1#2{%
\ifx\@captype\@IEEEtablestring%
  \parbox[t]{\hsize}{\footnotesize\noindent #1.~~ #2}%
  \@IEEEtablecaptionsepspace%
\else%
  \@IEEEfigurecaptionsepspace%
  \setbox\@tempboxa\hbox{\footnotesize #1.~~ #2}%
  \ifdim \wd\@tempboxa >\hsize%
    \setbox\@tempboxa\hbox{\footnotesize #1.~~ }%
    \parbox[t]{\hsize}{\footnotesize \noindent\unhbox\@tempboxa#2}%
  \else%
    \ifcenterfigcaptions \hbox to\hsize{\footnotesize\hfil\box\@tempboxa\hfil}%
    \else \hbox to\hsize{\footnotesize\box\@tempboxa\hfil}%
  \fi\fi\fi}
\makeatother

\title{\LARGE \bf
    Recompositional Robotics: Cross-Domain, Open-set, and \\
    Lifelong Modularity Beyond Morphology
}

\author{Steven Swanbeck, Jonathan Salfity, Corrie Van Sice, Robert Blake Anderson, and Mitch Pryor%
\thanks{The authors are with Texas Robotics and the Walker Department of Mechanical Engineering,
        The University of Texas at Austin, Austin, TX 78712, USA.}%
\thanks{This material is based upon work supported by the University Technology Development Directorate (UTDD), U.S. Army Futures Command (AFC), under Contract No. W911NF-24-C-0067. Any opinions, findings and conclusions or recommendations expressed in this material are those of the author(s) and do not necessarily reflect the views of the Army Futures Command, and no such endorsement should be inferred.\newline
DISTRIBUTION STATEMENT A. Approved for public release; distribution is unlimited. OPSEC \# 11117}
}

\begin{document}

\maketitle
\thispagestyle{empty}
\pagestyle{empty}


\begin{abstract}
    Research in modular robotics has produced capable approaches allowing a robot's morphology to change online, with recent efforts also developing approaches to decide which morphology to assume and automatically propagate that decision into the robot's motion planning and control.
    These approaches are powerful and increase adaptability in the field.
    However, an alternative objective is not to build robots whose \emph{structures} can change, but robots whose fundamental \emph{capabilities} can change, where capability is a joint function across several domains, including kino-dynamics, perception, compute, and high-level coordinating behaviors.
    A robot designed to be reconfigured across these domains has a greater capacity to alter its capability than one that can be reconfigured in a single domain.
    We refer to this cross-domain reconfigurability as \emph{integration span}, and recognize a complementary measure of the resistance to reconfiguration, which we refer to as \emph{integration inertia}.
    Current modular robots have reduced integration inertia in the structural domain while it remains high in the other domains that contribute to integration span.
    We assert that the systems that can provide the most utility through reconfiguration in practice are those maximizing span and minimizing inertia and call this general problem \emph{recompositional robotics}: adaptation over a heterogeneous set of modules including hardware, software, compute, and behavior that abstracts each component by the interfaces it requires and provides such that they can be reasoned over holistically.
    We define the problem, ground it in two deployed systems and active research efforts, and pose open questions about the future of recompositional robotics.
\end{abstract}
\section{Background and Motivation}
\label{sec:motivation}


General-purpose robots are trending.
For example, behavior generation for humanoids is advancing quickly on the presumption that one sufficiently capable body driven by a sufficiently general policy can cover a sufficiently broad distribution of useful tasks.
Yet, even these systems inherit the rigidity of the hardware and software subsystems underlying their policies, which remain effort-intensive to adapt when tasks or environments change.
Reconfigurable robots approach this problem differently; instead of a single body and set of functionalities that are adequate everywhere, a robot should instead alter its capabilities to specialize for the task at hand while remaining flexible over the tasks that follow \cite{trends_in_reconfigurable_modular_robots_2017,modular_reconfigurable_review_2019,modular_reconfigurable_review_2025}.
While this is an intuitive approach for many practical applications of autonomy, modular products remain largely a research and not industrial solution except for occasional instances of modular tooling. 
This is largely because modular solutions have yet to provide complete solutions in terms of \emph{capabilities}. 
\emph{Developing} new capabilities is of interest in research but \emph{deploying} new capabilities is what is needed in industry.

The field of modular and reconfigurable robotics has spent decades developing impressive solutions to allow autonomous systems to change structure, including self-reconfigurable systems that rearrange near-identical atoms into target morphologies \cite{modular_grand_challenges_2007,trends_in_reconfigurable_modular_robots_2017,survey_of_reconfigurable_mechanisms_design_2024}, aerial modules that join and cooperate midair \cite{self_assembly_swarm_2014,modquad_2018}, computationally designed organisms assembled from simpler parts \cite{reconfigurable_organisms_2020}, and field platforms whose structure is modified to suit the job at hand \cite{concert_2026,hefty_2024}.
Many modular systems can recognize when their own topologies change, regenerate kinematic, dynamic, and geometric models from data carried by mated modules, and be operational again in minutes \cite{concert_2026}, and some approaches can even be provided a task and asked which feasible morphology best suits it \cite{design_and_pose_optimization_2024}.

\newcommand{\ymark}{$\checkmark$}
\newcommand{\pmark}{$\circ$}
\newcommand{\nmark}{$\times$}
\newcommand{\qtriple}[3]{#1\,#2\,#3}

\begin{table*}[t]
\centering
\caption{Positioning against neighboring threads in literature and work-to-date toward recompositional robots.
\ymark~supported, \nmark~not supported.
The final block is the two systems of \Cref{sec:wtd} and the long-term objective, corresponding to points 1, 2, and $\star$ of \Cref{fig:wtd}.
}
\label{tab:positioning}
\footnotesize
\setlength{\tabcolsep}{5pt}
\begin{tabular*}{\textwidth}{@{\extracolsep{\fill}}l l l c@{}}
\toprule
\textbf{Thread \& Representative Work}
& \textbf{Integration Span}
& \textbf{Integration Inertia}
& \textbf{Open-Set} \\
\midrule

Self-reconfigurable modular robots \cite{modular_grand_challenges_2007,trends_in_reconfigurable_modular_robots_2017,survey_of_reconfigurable_mechanisms_design_2024}
    & HW
    & Low for morphology only
    & \nmark \\

Task-driven modular platforms \cite{concert_2026,design_and_pose_optimization_2024,hefty_2024}
    & HW
    & Low for morphology and kino-dynamic model only
    & \nmark \\

Plug-and-produce \cite{plug_and_produce_collaborative_2017,generic_plug_and_produce_2021,plug_and_produce_review_2024}
    & HW
    & Low for registered and preconfigured devices only
    & \nmark \\

Self-adaptive software \cite{mros_2022,software_reconfiguration_2025,temoto_2022}
    & SW
    & Low within variant space
    & \nmark \\

Task and behavior planning \cite{tamp_review_2021,plansys2_2021,skiros2_2023}
    & Behavior
    & Low for top-level plan generation only
    & \nmark \\
Monotone co-design \cite{co_design_theory_2015,co_design_embodied_intelligence_2021,codei_2025}
    & HW, SW, compute
    & High, requires offline re-modeling
    & \nmark \\

\midrule

Unseen payload integration \cite{deployment_is_not_destiny_2026}
    & HW, SW, compute
    & Low, minutes w/ no manual intervention
    & \ymark \\

Holistic self-recomposition \cite{lifelong_recomposition_2026}
    & HW, SW, behavior
    & Low within closed set of modules
    & \nmark \\

\textbf{Recompositional robotics ($\star$)}
    & \textbf{HW, SW, compute, behavior}
    & \textbf{Low across all domains}
    & \ymark \\
\bottomrule
\end{tabular*}
\end{table*}

These efforts point in the right direction: a system whose morphology remains mutable at runtime leaves free a set of design variables that a fixed-body robot commits to at build time. 
The value is not that a larger configuration space admits a larger capability space (such a claim is trivial), but rather that variables left free can be resolved against the actual task and conditions the robot encounters during operation rather than those anticipated by the designer.

However, a robot's capability is a joint function of its structure, kino-dynamics, perception, compute, and intelligence, not of morphology alone. Reconfiguration methods confined to structural morphology therefore realize only a fraction of the available gains, leaving the analogous gains in the other domains untouched. 
Further, when a change in the robot's task, environment, or intrinsic state renders its current configuration unfit, the least costly restoration may lie outside morphology, or may require coordinated change across several domains. Neither is visible to a system that treats morphology as the only reconfigurable domain.
Further, existing modular systems make reconfiguration tractable by fixing the catalog of modules, pre-installing required software on the host, or pre-authoring dependencies and transition rules \cite{plug_and_produce_collaborative_2017,generic_plug_and_produce_2021,plug_and_produce_review_2024,concert_2026}.
This reconfiguration using a closed set of modules created at design-time means adaptability during runtime remains bounded by the foresight of the designers, and risks rendering even modular systems ineffective when truly unforeseen requirements emerge.

We refer to the continuous ability to reconfigure across subsystem domains as \emph{integration span} and claim that a system that maximizes integration span, by nature of having mutability across a wide range of subsystems that contribute to its capability, will be most successful at altering its capability when required during operation.
However, reconfigurability is not binary, as required time, effort, and autonomy vary significantly between systems.
Thus, we also consider the \emph{integration inertia}, or resistance of a system to reconfiguration.
Using these two measures, a traditionally integrated robot, such as an industrial robot in a manufacturing setting, has low span and high inertia, keeping only task planning mutable during deployment (with possible occasional software updates) and requiring significant manual intervention to meaningfully alter its hardware, software, or compute \cite{component_based_robot_engineering_p1_2009,software_reconfiguration_2025}.
We also contrast the closed-set nature of existing reconfigurable systems by defining \emph{open-set} reconfiguration, which enables a system to admit modules that were not known by the robot in advance and that may not have even existed when it was first deployed.

The primary claim of this abstract is that reconfiguration should be extended to consider holistic capability beyond morphology, and that we need not only mechanisms to support this expanded notion of reconfiguration but also methods to reason over and generate cross-domain system configurations that can be actualized by a robot during operation.
\Cref{sec:recomposition} frames this problem and \Cref{sec:wtd} grounds it in recent efforts and current research directions before \Cref{sec:fw} concludes with directions for future work and open questions.
\section{Recompositional Robotics}
\label{sec:recomposition}

We use the term \emph{recompositional} to describe our setting rather than using \emph{modular} or \emph{reconfigurable}, which the existing literature tightly associates with morphological reconfiguration.

\begin{quote}\itshape
    A recompositional system adapts by re-deriving a joint configuration over a heterogeneous set of parts including electromechanical modules, software components, compute resources, and behaviors, each described only by the interfaces it requires and provides.
    Composition is the assembly of parts along these interfaces and adaptation is the synthesis and adoption of a new composition.
\end{quote}

Ranging over a heterogeneous set of parts increases integration span, while describing every part only by its interfaces lowers integration inertia. 
A part known solely by what it requires and provides can be implemented or substituted locally in a larger composition without re-deriving anything with which it does not interact.
Neither property is sufficient alone; a formulation may span many domains but synthesize new compositions too slowly for real-time deployment, and one that can synthesize a composition instantly over only one or two domains may discover a suboptimal solution if it discovers one at all.
We emphasize three features that distinguish this setting, and \Cref{tab:positioning} summarizes its positioning.

\textbf{1) A capability model that spans domains.}
In existing self-reconfigurable modular robotics, the configuration free variables include the physical structure of the system, including the selection of modules and their relative positions and posture.
In our broader definition of a recompositional system, the free variables still include the physical structure and also the running software modules and their parameterization, the allocation of available compute, and the sequence of actions performed by the resulting system.
The practical consequence of this view is that the minimal restoration to an unfit configuration is often not purely morphological or not morphological at all.
For example, a vision sensor that becomes ineffective in low-light conditions might require different drivers, offloading inference to a peer with superior capabilities that can run a more powerful image processing model, or a re-parameterization of a behavior, all of which may be faster, less effort-intensive, or safer than a hardware update.
A formulation that only spans morphology cannot see those alternatives, let alone compare them to find a cross-domain optimum.
Often, no single domain suffices.
For example, a newly mounted sensor is worthless without software that consumes it and a behavior to act on what that software produces, and that coupling is precisely what a per-domain approach cannot express.

\textbf{2) An open vocabulary, not a closed set of modules.}
Closing the set of available modules makes reconfiguration tractable, but it bounds runtime adaptation to design-time assumptions.
We are interested in the case where a system can utilize modules that are genuinely \emph{unseen}, meaning it has no prior knowledge of what they are, how to use them, or what they afford before they are integrated.
This property allows a robot to utilize resources for which no need was anticipated prior to its deployment.
Furthermore, this same property allows a robot deployed today to quickly integrate a module developed tomorrow with technology or capabilities that did not even exist at the time of its original deployment, all without a return to a workshop, factory, or otherwise being removed from its mission.
This is not an exotic demand in the contexts for which modular systems are often proposed: disaster response, search-and-rescue, inspection, and construction frequently surface requirements the robot was not equipped for, where the capability may exist but cannot be integrated within the mission's time budget.

\textbf{3) Lifelong queries, not just a single synthesis.} 
A configuration synthesizer receives only an objective and reasons over the set of available modules to produce a configuration that achieves that objective.
A long-lived system must instead introspect its own persistent configuration repeatedly, and must produce answers to at least four distinct questions. 
\textbf{Q1}: Am I still fit? 
\textbf{Q2}: If not, why? Which requirements or conditions are violated? 
\textbf{Q3}: What is the minimal restoration with respect to time, effort, cost, or safety, and how much must the incumbent configuration change to achieve it? 
\textbf{Q4}: Is that restoration worth its cost, or should I continue in a suboptimal but operational configuration?

Q1 makes adaptation autonomous rather than commanded.
Q2 makes it legible to a human teammate and helps guide restoration. 
Q3 and Q4 cannot be expressed at all by a synthesis-only formulation. 
Reconfiguration is never free, so the choice between paying for the optimum and persisting with a degraded configuration is itself a decision that depends on the system, task, and environment, and therefore should be handled at runtime rather than being fixed in advance by a designer.

\section{Work To Date}
\label{sec:wtd}

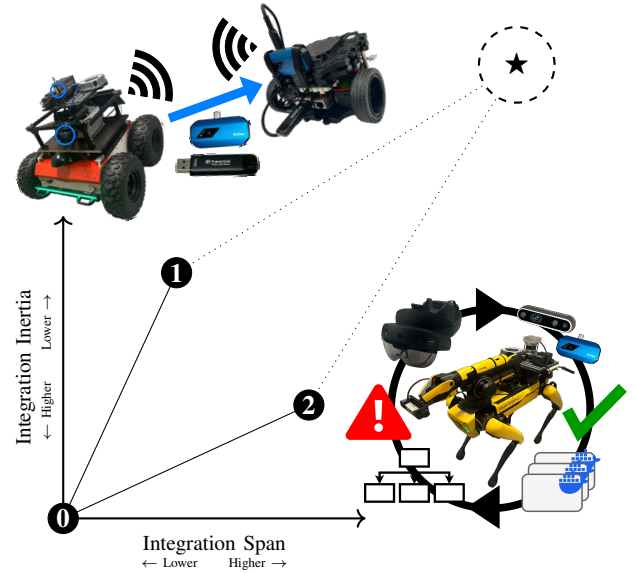
\begin{figure}
\begin{tikzpicture}[scale=0.5]
    \pgfdeclarelayer{background}
    \pgfsetlayers{background,main}
        
    \draw[->, thick] (0,0) -- (8,0) node[right] {};
    \draw[->, thick] (0,0) -- (0,8) node[above] {};
    

    
    \node[rotate=90, above] at (-0.5,4) {\footnotesize Integration Inertia};
    \node[rotate=90, above] at (0,4) {\tiny $\leftarrow$ Higher \qquad Lower $\rightarrow$};
    \node[below] at (4,-0.2) {\footnotesize Integration Span};
    \node[below] at (4,-0.8) {\tiny $\leftarrow$ Lower \qquad Higher $\rightarrow$};

    \coordinate (base) at (0,0);
    \node[circle, fill=black, draw=black, text=white, font=\bfseries, inner sep=1pt]
          (objnode) at (base) {0};

    \coordinate (dind) at (3,6.5);
    \node[circle, fill=black, draw=black, text=white, font=\bfseries, inner sep=1pt]
          (objnode) at (dind) {1};
    
    \coordinate (lrr) at (6.5,3);
    \node[circle, fill=black, draw=black, text=white, font=\bfseries, inner sep=1pt]
          (objnode) at (lrr) {2};

    \coordinate (obj) at (12,12);
    \node[circle, fill=white, text=black, font=\bfseries, draw=black, dashed, thick, minimum size=1cm, inner sep=1pt] at (obj) {$\bigstar$};

    \node[rotate=20] (dind_img) at (3.5,10.75) {\includegraphics[height=2.2cm]{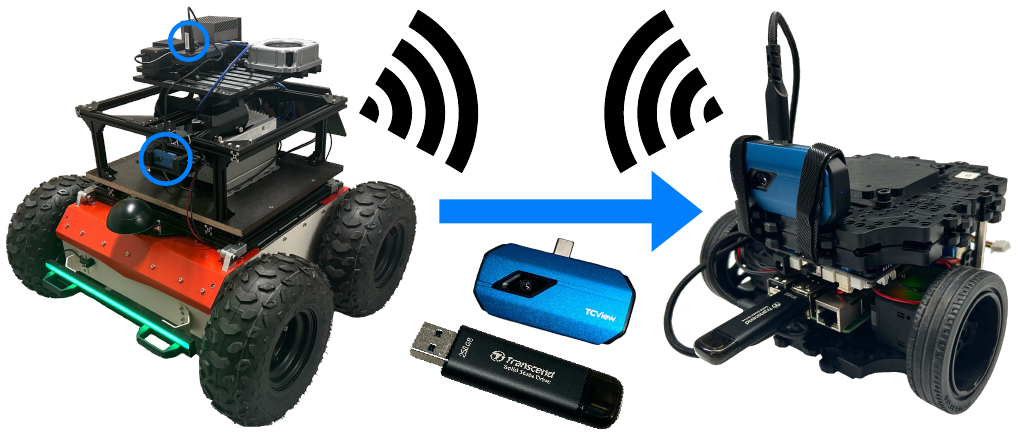}};

    \node[rotate=-0] (lrr_img) at (11.25,2.9) {\includegraphics[height=3.2cm]{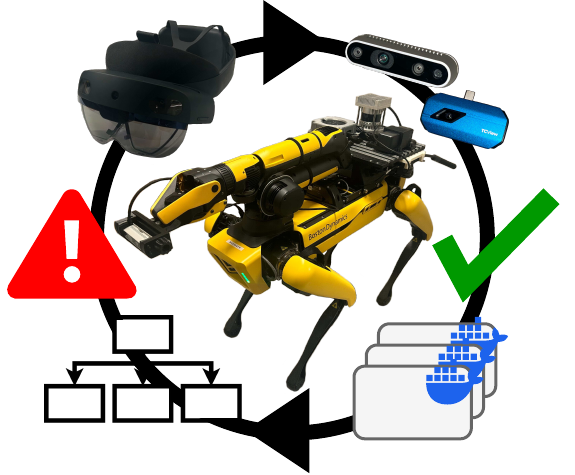}};


    \begin{pgfonlayer}{background}
        \draw[thin,color=black] (base) -- (lrr);
        \draw[thin,color=black] (base) -- (dind);
        \draw[thin,dotted,color=black] (lrr) -- (obj);
        \draw[thin,dotted,color=black] (dind) -- (obj);
    \end{pgfonlayer}
\end{tikzpicture}
\caption{%
    Integration span versus integration inertia.
    0) A system that is not designed to reconfigure has maximum inertia and minimum span.
    1) By enabling plug-and-play integration with no prior knowledge, support for hardware, software, and compute modules, and resource sharing in distributed systems, \cite{deployment_is_not_destiny_2026} significantly lowers the integration inertia of deployed robots.
    2) By enabling a robot to simultaneously reason over its deployed hardware, software, and behavior, \cite{lifelong_recomposition_2026} supports cross-domain recomposition that can respond dynamically as the robot, its task, or its environment change. This work significantly improves system-level reasoning and autonomy, but does not demonstrate an open set of modules.
    $\star$) The north star for recompositional robotics. A system with maximum span and minimum inertia has the ability to reconfigure all aspects of itself in real-time to respond to emergent requirements.
}
\label{fig:wtd}
\end{figure}

We ground this problem in two recent efforts \cite{deployment_is_not_destiny_2026,lifelong_recomposition_2026} that each push primarily along one axis of the span vs. inertia problem while conceding the other (\Cref{fig:wtd}).
We state what each work does rather than how it works, with full details left to the corresponding manuscripts. 

\subsection{Lowering Inertia with an Open Set of Modules}
\label{subsec:wtd:fraps}

The first system relaxes the assumption of a closed set of modules by providing abstractions and a corresponding framework that allow new hardware, software, and compute payloads to be introduced to a robot during operation and integrated into its task planning and execution within minutes \cite{deployment_is_not_destiny_2026}.
Hardware connects via standard USB interfaces, with associated drivers and hardware-independent software resources supplied on the connected drives.
Distributed peers and compute payloads can also be discovered over Ethernet or wireless connections, and capabilities afforded by one system become accessible to its distributed peers. 
None of this requires a developer on site, a reboot, or the restart of any system processes.

We evaluated the system in two field exercises. 
In a disaster response exercise at an operational nuclear reactor facility, a quadruped's payload suite was recomposed twice mid-mission in \SI{2}{\minute}\,\SI{16}{\second} and \SI{3}{\minute}\,\SI{19}{\second} to continue operation when emergent mission needs rendered the robot incapable of performing an intended task.
In a search-and-rescue scenario, an unspecialized mobile robot with limited compute resources was first configured for the task in \SI{7}{\minute}\,\SI{22}{\second} with a distributed peer providing superior compute resources. 
Once a narrow entryway to an affected building was discovered, payloads were swapped to a smaller robot that also leveraged the capabilities of the same distributed peer, reconfiguring it in \SI{5}{\minute}\,\SI{39}{\second} to complete the mission.
These reconfiguration processes could have taken hours of expert effort without plug-and-play cross-domain payloads, far outside the affordable response window in sensitive applications.

\subsection{Extending Span with Self-Driven Cross-Domain Recomposition}
\label{subsec:wtd:lifelong}

The first system exercises no judgment about why, when, or how it should reconfigure; while it enables rapid recomposition using unseen and heterogeneous modules, a human teammate ultimately drives the recomposition process. 
The second work supplies that judgment, enabling the robot to have agency over its own recomposition.
It introduces an abstraction that models hardware, software, and behavior modules uniformly and reasons over them together, carrying co-design theory past the design stage and into runtime \cite{lifelong_recomposition_2026}. 
Recomposition is solved as a satisfiability problem that persists the composition as a queryable logical model rather than a one-shot synthesis result, and can therefore answer the diagnostic questions raised in \Cref{sec:recomposition} rather than merely producing a standalone composition.
In a search-and-rescue deployment, the robot recognized that it was unfit on three occasions, each for a distinct reason. 
Each time it re-synthesized a new composition spanning hardware, software, and behavior in under \SI{7}{\second} onboard. 
It then enacted the software changes itself, regenerated its own behavior tree, and issued the hardware changes as instructions to a human teammate wearing an augmented-reality headset.
When faced with mission-time perturbations representative of a long-lived reconfigurable system, the approach recovered an optimal configuration with zero manual model edits. 
In contrast, optimal-numeric, stream-based, and satisficing SMT planners that were evaluated against required 22--34 hand edits to their models to answer each of these questions, limiting the realizable autonomy of these approaches to the recompositional robotics setting.

\section{Future Work and Open Questions}
\label{sec:fw}

The works to date are limited in complementary directions: \cite{deployment_is_not_destiny_2026} integrates modules it has never seen but exercises no judgment about when or how it should reconfigure, while \cite{lifelong_recomposition_2026} provides that judgment but only over a closed set of modules known in advance.
Unifying these approaches is our current objective, to both enable a system to reason over its own fitness throughout a mission and dynamically incorporate new resources it has never seen before during operation.

Unifying these approaches raises at least four open problems.
First, the state representation maintained in \cite{lifelong_recomposition_2026} is not relational, and therefore struggles to express dependencies between resources within the distributed contexts prioritized in \cite{deployment_is_not_destiny_2026}.
Second, its satisfiability-based synthesis approach assumes a finite, enumerable universe of resources and modules, in tension with the open-set property that motivates this work.
Third, extending synthesis from a single robot to a larger network of systems introduces contention over shared resources and the question of centralized versus federated solving.
Fourth, domains carry different costs of reconfiguration, and reasoning over all of them identically at every timestep risks being prohibitively expensive as the frequency of reevaluation and recomposition increases.
The domain most likely to require an increased frequency is task planning; \cite{deployment_is_not_destiny_2026} and \cite{lifelong_recomposition_2026} both generate longer-horizon behavior trees that will execute on the timescale of seconds or minutes at comparable frequency to the other recomposable domains, but an alternative approach may be to plan shorter-horizon action trajectories at a higher frequency.

These problems are not an exhaustive set, and we expect resolving them to surface further open questions as recompositional systems are designed and deployed more broadly.


\printbibliography

\end{document}